\documentclass[sigconf,nonacm,10pt]{_sty/acmart}
\usepackage{comment}
\usepackage{multirow}
\usepackage{colortbl}
\usepackage{lipsum}
\usepackage{balance}
\definecolor{mygray}{gray}{.9}
\definecolor{mygray95}{gray}{.80
}
\AtBeginDocument{%
  \providecommand\BibTeX{{%
    \normalfont B\kern-0.5em{\scshape i\kern-0.25em b}\kern-0.8em\TeX}}}

\setcopyright{acmlicensed}
\copyrightyear{2018}
\acmYear{2018}
\acmDOI{XXXXXXX.XXXXXXX}

\begin{document}

\title[QuadraNet V2: Efficient and Sustainable Training of High-Order Neural Networks with Quadratic Adaptation]{\huge{QuadraNet V2: Efficient and Sustainable Training of High-Order Neural Networks with Quadratic Adaptation}}

\author{\Large{Chenhui Xu$^1$, Xinyao Wang$^2$, Fuxun Yu$^3$, Jinjun Xiong$^{1,*}$, Xiang Chen$^{4,*}$}}
\author{\large{$^1$University at Buffalo  $^2$Dalian University of Technology $^3$Microsoft  $^4$Peking University}}
\author{\Large{}}
\renewcommand{\shortauthors}{Chenhui Xu, et al.}

\begin{abstract}
    %Deep learning models are extensively utilized for downstream tasks, typically through pre-training on large-scale datasets followed by fine-tuning on specific task datasets. However, the emergence of large foundation models with billions of parameters has made full fine-tuning at such scales impractical. To address this, researchers have explored adapting specific parameters or integrating supplementary adapters. Traditional adapters, though, often oversimplify the distribution shift between pretraining and task data, leading to suboptimal performance. This paper proposes a solution in the form of QuadraNet V2, leveraging quadratic kernel adapters to model high-order relations and capture nuanced data shifts effectively. QuadraNet V2 highlight the efficiency  gains enabled by the quadratic adapters and performance gains achieved by exploring more complex non-linear space with kernel-enhance. Through empirical evaluation, we demonstrate the superior performance of \colorbox{gray} across various downstream tasks compared to vanilla QuadraNet, underscoring the efficacy of quadratic adapters in addressing adapting challenges in deep learning models. 

Machine learning is evolving towards high-order models that necessitate pre-training on extensive datasets, a process associated with significant overheads. Traditional models, despite having pre-trained weights, are becoming obsolete due to architectural differences that obstruct the effective transfer and initialization of these weights. To address these challenges, we introduce a novel framework, QuadraNet V2, which leverages quadratic neural networks to create efficient and sustainable high-order learning models. Our method initializes the primary term of the quadratic neuron using a standard neural network, while the quadratic term is employed to adaptively enhance the learning of data non-linearity or shifts. This integration of pre-trained primary terms with quadratic terms, which possess advanced modeling capabilities, significantly augments the information characterization capacity of the high-order network. By utilizing existing pre-trained weights, QuadraNet V2 reduces the required GPU hours for training by 90\% to 98.4\% compared to training from scratch, demonstrating both efficiency and effectiveness.

\end{abstract}

\maketitle
\let\thefootnote\relax\footnotetext{*Corresponding Authors. <jinjun@buffalo.edu><xiang.chen@pku.edu.cn>}
\section{Introduction}
For contemporary research on deep neural network (DNN) models, there is a pronounced shift towards the deployment of increasingly large-scale models. These models generally require extensive pre-training on massive datasets to perform effectively~\cite{touvron2023llama,dosovitskiy2020image}. For example, the training of a Vision Transformer~\cite{dosovitskiy2020image} on ImageNet-21K consumes around 10,000 GPU hours on advanced Nvidia A100 GPUs. The financial implications of such pre-training are substantial, often amounting to hundreds of thousands of dollars. This burgeoning reliance on large-scale pre-training DNN models not only raises significant sustainability issues, but also limits the practical wide application of DNN models due to the high costs and computational demands involved.

\begin{figure}[t]
  \centering
  \includegraphics[width=\linewidth]{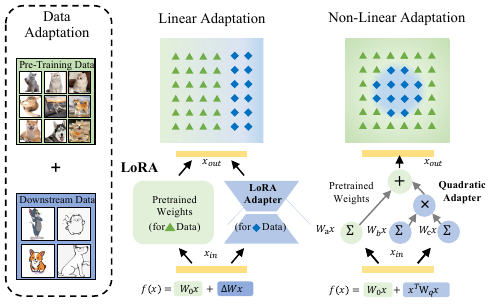}
   \vspace{-5mm}
  \caption{Linear v.s. Nonlinear Data Adaptation}
  \label{fig:shift}
  \vspace{-6mm}
\end{figure}

To circumvent the inefficiencies of repetitive pre-training, there has been a trend toward utilizing adaptation techniques to reuse those pre-trained linear DNN models for downstream tasks. These methods, such as those pioneered by the Low-Rank Adaptation (LoRA)~\cite{hu2021lora} approach, aim to fine-tune pre-existing models for new tasks without the need for complete retraining by adding an adaptation matrix on the original weights. 
%However, these techniques primarily address linear discrepancies between the pre-training and target task distributions. 
For example, as shown in Fig.\ref{fig:shift}, LoRA~\cite{hu2021lora} adopts a linear transformation matrix $\Delta W$ to model the distribution shift of the inputs from the pre-training data to the downstream data.
%\jx{This is }

In practice, the distribution shift of the input data is not always linear.
To address the complex non-linear data issues, there has been growing interest in developing high-order DNNs models, such as HorNet~\cite{rao2022hornet}, MogaNet~\cite{li2024moganet}, and QuadraNet~\cite{fan2023expressivity,xu2023quadranet,xu2022quadralib}. These networks are designed to more accurately model the intricate nonlinear relationships within data, offering a promising avenue for enhancing the model's capability. 
However, training these high-order DNNs is even more complicated than training the traditional linear DNNs. To make it even worse, the LoRA-like adaptation techniques would not work well to fine-tune toward a pre-trained high-order DNN for downstream tasks.
This is due to the nature of LoRA's linear adaptive matrix, which can only model the distribution shift in a linear fashion, but not the more complex, nonlinear shifts for real-world applications.

 To address this challenge, among many existing high-order DNN models, we note that the high-order interactions in QuadraNet~\cite{fan2023expressivity,xu2023quadranet,xu2022quadralib}
 have some unique characteristics. Different from other high-order DNN models, QuadraNet's high-order interactions 
 are embedded inside the neurons. This has made quadratic DNNs an architecture-agnostic high-order DNN. In other words, it is relatively easy to transform any existing linear DNN model to a quadratic one by replacing the linear neurons with their quadratic counterpart as shown in the work of QuadraLib~\cite{xu2022quadralib}.
This property helps us, at a high level, to treat a quadratic DNN consisting of two parts: a DNN  architecture that is the same as any linear DNN architecture, and the selective application of quadratic neurons. This insight is significant because a quadratic neuron is represented by the addition of two terms: a linear term and a quadratic term. By way of analogy to LoRA's adaptation formula as shown in Fig.\ref{fig:shift}, we can treat QuadraNet's linear term the same as
the pre-trained weights from a linear
DNN, while
the quadratic term as an adaptation
a term that upgrades a linear DNN
to a high-order QuadraNet.
This insight makes it possible to train completely new high-order neural networks starting from existing pre-trained linear DNNs while achieving adaptation to effectively capture non-linear distribution shifts towards downstream data.

 In this paper, we introduce QuadraNet V2, an innovative, efficient, and sustainable framework for training high-order neural networks. QuadraNet V2 initializes the linear term of the quadratic neuron using a standard neural network architecture, subsequently employing the quadratic term to adaptively learn the non-linear distributional shifts. This model capitalizes on the synergistic and interactive effects of the high-order and primary terms within the quadratic neurons, both in terms of information processing and computational independence. This offers a straightforward, modular, and efficient high-order network training paradigm that effectively leverages pre-existing pre-trained assets. To enhance the efficiency of the quadratic neurons, we employ a low-rank and atrous design, which optimizes the modeling of high-order interactions within the neuron's receptive field, minimizing computational overhead. Our experimental results demonstrate that QuadraNet V2 can reduce GPU training time by up to 98.4\% compared to training from scratch, thereby underscoring its potential as a sustainable, efficient, and effective architecture for next-generation high-order neural models.

\textbf{Contributions.} We make the following contributions:
\begin{itemize}
\item \textbf{Sustainable Training Framework.} We propose QuadraNet V2, a groundbreaking framework for the efficient and sustainable training of high-order neural networks that effectively utilize legacy pre-trained weights, offering fresh perspectives on deriving value from existing pre-trained neural network models.

\item \textbf{Neural-level Adaptation.} We reveal the impact of high-order interaction on the model's proficiency in adapting to nonlinear distribution shifts from pre-trained to downstream data. We detach high-order interactions from the architectural constraints of the model with the introduction of quadratic neurons, enabling the construction of high-order networks through a novel quadratic adaptation approach.

\item \textbf{Efficient Quadratic Design.} We optimize the design of the quadratic network for improved computational efficiency through innovative low-rank and atrous design of the quadratic terms and integration of external acceleration mechanisms.
\end{itemize}

\begin{figure}[t]
  \centering
  \includegraphics[width=\linewidth]{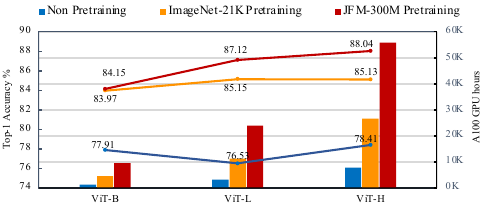}
  \vspace{-8mm}
  \caption{Model performance on ImageNet-1K and GPU time required for different scales of pre-training. }
  \label{fig:pretraining}
    \vspace{-6mm}
\end{figure}

\section{Theoretical Analysis}

\subsection{Pre-Training: Where Are We Today?}
\begin{comment}
Pre-training is considered a key element in today's neural network construction.
Pre-training a model on a large, diverse dataset improves learning performance and generalization, enabling it to perform better on specific tasks with less task-specific data. This technique enhances feature extraction, reduces biases, and facilitates advanced applications like zero-shot and few-shot learning. As shown in Fig.~\ref{fig:pretraining}, as the scale of the model increases, the performance gains from pre-training on larger-scale datasets are significant. However, as the scale of pre-training increases, the overhead of the training process increases dramatically.

As shown in Figure 2, pre-training a ViT-H model on A100 with 7 epochs of JFM-300M would take about 56,000 GPU hours. Such an arithmetic requirement would cost about \$100,000 at the current time (April 2024). However, this is not the limit of pre-training, as today's large-scale models are often pre-trained using web-wide data, which can cost millions of dollars at this scale of pre-training. The problem is that every time we build a new model structure, we need to start pre-training the model from scratch. This means that old models that have been pre-trained at great expense will be completely abandoned, which is a huge waste of resources. In this regard, we pose the question: 

\textit{Can we utilize the remaining value of the existing legacy trained models when building a new more powerful models?}
\end{comment}

Pre-training is recognized as a crucial component in the construction of modern neural networks. By pre-training a model on a large dataset, its learning performance and generalization capabilities are significantly enhanced, thereby allowing it to perform better on specific tasks using less task-specific data. This method not only improves feature extraction but also minimizes biases and supports advanced applications such as zero-shot and few-shot learning. As depicted in Fig.~\ref{fig:pretraining}, the benefits of pre-training become more pronounced with increases in model scale, with substantial performance gains observed when using larger datasets. \textbf{Therefore, the machine learning community has now trained numerous models with large-scale pre-training.}

However, this scaling up of pre-training also leads to a dramatic rise in the training overhead.
As demonstrated in Fig.~\ref{fig:pretraining}, pre-training a ViT-H model on NVIDIA A100 GPUs across 7 epochs with the JFM-300M dataset would require approximately 56,000 GPU hours, translating to an estimated cost of \$100,000 as of April 2024. Yet, this is not the upper limit for pre-training costs. Contemporary large-scale models are often trained using web-scale data, potentially incurring expenses in the millions of dollars for such levels of pre-training. A significant issue arises because each new model architecture necessitates initiating pre-training from scratch, leading to the \textbf{abandonment of previously trained models}, which have been developed at considerable expense. This represents a substantial waste of resources.
In light of this, we pose the question:

\textit{Can the residual value of legacy trained models be utilized when constructing new, more powerful models?}

The primary motivation of this work is: We identify the opportunity to utilize the massively pre-trained weights that are available now to build a new generation of more powerful models with minimal training cost. Despite this opportunity, we still find the following challenges:

(1) There is still a distributed gap between pre-training data and downstream data, and picking an appropriate incremental approach to bridging them is difficult. (Section~\ref{sec:shift})

(2) Recent performant models are architecturally completely different from the pre-training weights that are already available, and such weights are completely meaningless for training the new models. (Section~\ref{sec:archi})

\begin{figure}[t]
  \centering
  \includegraphics[width=\linewidth]{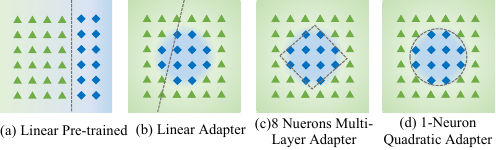}
  \caption{High-order Adaptation Capacity.}
  \label{fig:adapter}
  \vspace{-1mm}
\end{figure}

%Middle-scale Pre-training: Training on ImageNet-21K for 90 epochs. Large-scale Pre-training: Training on JFM-300M for 7 epochs. GPU hours are counted on Nvidia A100 80GB

%Observation 2 illustrates that the quadratic term of a quadratic neural network can be trained after the primary term. And now the AI community exists by a large number of ordinary neural networks pre-trained on large-scale datasets. Adding a quadratic term to these outdated neural networks can allow the network to be tuned into a more expressive quadratic neural network, giving the opportunity to form high-performance quadratic neural networks by adding a quadratic adapter without having to be pre-trained again on a mega dataset. But despite the fact that the representation ability of the quadratic adapters is significantly better than that of the regular adapters, as described in Observation 1, adding the quadratic adapters to all of the neurons in the entire network will still incur an exponential computational overhead. Unreasonable configuration of the quadratic adapter will bring huge computational overhead, which poses a challenge for the application of the quadratic adapter. Fortunately, quadratic adapters with high representation capacity density can be added to the network in small amounts.

\subsection{Difficulty of Modeling Non-linear Shift}
\label{sec:shift}
Pre-training data has a different distribution than downstream data. In response to this discrepancy, the dominant approach in the past has been the full fine-tuning of models based on the theory of transfer learning. Assume that we change the entire weight matrix to $W_{\text{Tuned}}$ by pre-training the weights $W$, with full fine-tuning of the training. But this approach faces the irreversibility of fine-tuning and huge computational overhead as the number of trainable parameters in $W_{\textit{Tuned}}$ equals $W$. A recent solution idea is to add a low-rank linear adaptation~\cite{hu2021lora} $\Delta W$ to the original weight $W$, so the tuned model would be $W+\Delta W$, where $\Delta W$ contains much fewer parameters compared with $W$. 

However as shown in Fig.~\ref{fig:shift}, the shift of data distribution often exhibits nonlinearity in the real world. Existing adapters tend to be linear transformations between states, which leads to the inability of adapters to model such nonlinear shifts with single-layer linear adapters, as illustrated in Fig.~\ref{fig:adapter}(b). The modeling of this nonlinearity requires a non-linear adapter, and the structure of such a multilayer adapter requires multiple basic neurons to perform. The multilayer adapter performs a piecewise linear fit. As shown in Fig.~\ref{fig:adapter}(c), the adapter consisting of 8 neurons does a correct classification of the variations of the tuning dataset, however, this piecewise linear modeling is obviously still sub-optimal.

\textbf{High-order architecture is a candidate for non-linear modeling.} High-order neural networks tend to have a greater ability to model nonlinearity in space, due to the high-order relationships of the data modeled in their models. Such high-order interaction mechanisms can be formulated as: 
\begin{equation}
	y_{\text{high-order}}^{(i,c)} = \sum_{j \in \Omega_i}\sum_{c'=1}^Cg(x)_{ij}T^{(c',c)}x^{(j,c')},
	\label{equ:highorder}
\end{equation}
where $\Omega_i$ corresponds to the mechanism's receptive field; $g(x)$ serves as an interactive weight matrix encapsulating the characteristics of the input; and $T^{(c',c)}$ reflects a transformation that operates channel-wise. The interaction between $x$ and the dynamically adjusted $g(x)$ results in complex, high-order neural interactions within the data. This high-order mechanism is natural for modeling nonlinear data. As illustrated in Figure \ref{fig:adapter}(d), we introduce a quadratic form as an adapter to the original classifier. This high-order model not only accurately classifies the sample points but also yields a smoother modeling of distributional shifts.

%\textbf{Observation 1} (Representation Capacity of Quadratic Form) The robust representational capabilities of a quadratic neural network primarily stem from its quadratic terms \cite{fan2023one}. Empirical evidence provided by Vanilla QuadraNet \cite{xu2023quadranet} substantiates that the quadratic form stands out as both computationally efficient and superior in performance for incorporating quadratic terms. As illustrated in Figure \ref{fig:adapter}(d), we introduce a quadratic form as an adapter to the original classifier. This adaptation not only accurately classifies the sample points but also yields a smoother and more effective modeling of distributional shifts.

%The strong representation capacity of a quadratic neural network comes mainly from its quadratic terms~\cite{fan2023one}. Vanilla QuadraNet~\cite{xu2023quadranet} empirically proves that the quadratic form is the computationally efficient and best performing form for constructing quadratic terms. As shown in Fig.~\ref{fig:adapter}(d), we add a quadratic type as adapter to the original classifier, which not only correctly classifies the sample points, but its also a smoother and better modeling of the distributional shifts.
%用一个toy case和 高维编码图片的解密/解码问题做一个例子
%\textbf{Observation 1} (Stage Training of QDNNs) 

%\textbf{Observation 2} (Representation of Quadratic Form)

\begin{figure}[t]
  \centering
  \includegraphics[width=\linewidth]{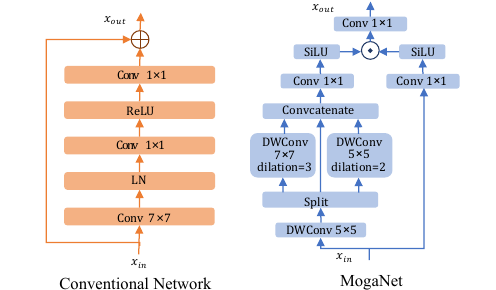}
  \caption{High-order models have different architecture with traditional neural networks.}
  \label{fig:archi}
  \vspace{-4mm}
\end{figure}

\subsection{Quadratic Net: Architecture-Agnostic High-Order Neural Interaction}

\textbf{Architectural Differences between High-Order and Traditional Models: }
\label{sec:archi}
\begin{comment}
Despite the strong nonlinear characterization capabilities of high-order models, such high-order is often defined by model architecture. The traditional model is fundamentally different from the high-order model in terms of architecture. As shown in the Fig~\ref{fig:archi}, the current state-of-the-art high-order network, MogaNet, has an extremely complex structure for building high-order interactions, which is significantly different from the concise traditional neural networks. This leads to the inherent difficulty of initializing such a high-order network with our existing pre-trained weights.

Quadratic neural networks reduce the granularity of high-order interaction to the neuron level, which means that we can build a architecture-independent neural network for high-order interaction. A quadratic neuron has the form:

\begin{align}
      y = & \sigma(X^TW_QX+W_CX+b)\notag \\
     \label{equ:neu}
\end{align}

where $W_Q \in \mathbb R^{n \times n}$ is a parameter matrix for quadratic term with rank $k$. 
We can use such neurons to build neural networks that are identical in architecture to conventional neural networks.
\end{comment}
Despite the potent nonlinear characterization capabilities inherent in high-order models, their complexity is often dictated by the architectural design. In stark contrast to traditional models, high-order models exhibit fundamentally different architectural structures. As illustrated in Fig~\ref{fig:archi}, the state-of-the-art high-order network, MogaNet, features an intricate structure specifically engineered to facilitate complex high-order interactions, markedly diverging from the simplicity of standard neural networks. This complexity presents substantial challenges in initializing such high-order networks using existing pre-trained weights.

\textbf{Architecture-Agnostic High-Order Quadratic Neuron}: Quadratic neural networks simplify the implementation of high-order interactions by localizing these interactions to the neuron level. This approach enables the development of architecture-independent neural networks capable of high-order interactions. A typical quadratic neuron is defined as:

\begin{equation}
y =  \sigma(X^TW_QX+W_CX+b),
\label{equ:neu}
\end{equation}

where $W_Q \in \mathbb R^{n \times n}$ represents the parameter matrix for the quadratic term with rank $k$. Employing such neurons allows for the construction of neural networks that maintain the architectural identity of conventional neural networks, thereby simplifying the integration of high-order functionalities.

\subsection{Training QDNNs in Stages: Where We Are Going toward!}

Lifelong learning strategies, as outlined by Zenke et al. \cite{zenke2017continual}, suggest a staged approach to training model parameters. In this approach, certain parameters of the model are trained initially on a specific dataset. Once these parameters reach optimal performance on that dataset, they are then frozen, while the remaining parameters are trained on other datasets.
%Lifelong learning~\cite{zenke2017continual} gives tips that the parameters of the model are trainable in stages. Specifically, a model can first train some of its parameters on a certain size of dataset, and after reaching optimality on the dataset, freeze this part of the parameters and train the rest of the parameters on other datasets.

\begin{figure}[t]
  \centering
  \includegraphics[width=\linewidth]{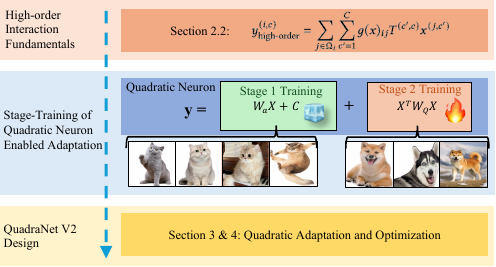}
  \caption{Stage Training of QDNNs.}
  \label{fig:neruon}
  \vspace{-4mm}
\end{figure}

\textbf{Observation} (QDNNs are trainable in stages.) 
%\begin{figure}[t]
  %\centering
  %\includegraphics[width=\linewidth]{_fig/block图.pdf}
  %\caption{叫啥来着等等改}
 % \label{fig:shift}
 % \vspace{-3mm}
%\end{figure}
As depicted in Fig.~\ref{fig:neruon}, the quadratic and primary terms of a quadratic neural network are linked via addition, forming two branches amenable to parallel computation. Employing the method outlined above, we first conduct one stage of training on the primary term, followed by freezing it and subsequently training the quadratic term on the remaining data. Through this process, we achieved performance nearly equivalent to training the quadratic neural network directly.

\textbf{Opportunities and Challenges}: Stage training observation highlights the feasibility of training the quadratic term after the primary term in a quadratic neural network. Adding a quadratic term as an adaptation to existing networks can enhance their expressiveness, creating high-performance quadratic neural networks through the addition of quadratic adapters \textbf{without retraining on large pre-training datasets}. However, the indiscriminate addition of quadratic adapters to all neurons leads to significant computational overhead. Improper configuration exacerbates this issue, posing challenges for practical implementation. Fortunately, the judicious incorporation of quadratic adapters in small increments can mitigate these challenges.

\begin{figure*}[t]
  \centering
  \includegraphics[width=\linewidth]{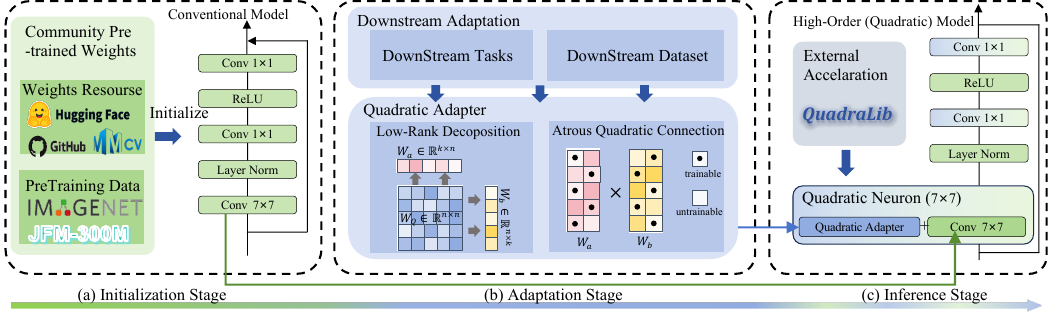}
  \caption{Overview of QuadraNet V2: (a) Directly pull existing model architecture and its weights trained on the pre-training data from community for model initialization. (b) A performant Quadratic Adapter with efficient Low-Rank Decomposition and Atrous Design. (c) Inference as library-accelerated quadratic neural network.}
  \Description{A woman and a girl in white dresses sit in an open car.}
  \label{fig:overview}
\end{figure*}

%第6栏末尾

\section{Design Methodology}

\subsection{QuadraNet V2 Overview}

Fig.~\ref{fig:overview} illustrates the overview of our proposed QuadraNet V2 framework, which consists of the following flow scheme:

(a) Pre-training/Initializing an ordinary neural network: We directly use existing convolutional neural network models that have been pre-trained on large datasets for initialization.

(b) Adaptation with Quadratic Adapter: a quadratic term for model tuning, with a low-rank decomposition to the quadratic term, an atrous design that sparsifies the quadratic adapter for a performant high-order adaptation of the model.

(c) Inference as accelerated Quadratic Neural Networks: With the incorporation of a quadratic adapter, the network can reason as a whole as a quadratic neural network that has been deeply accelerated by the quadratic neuron computational library (QuadraLib~\cite{xu2022quadralib}).

We take a case study of our QuadraNet V2 framework on a pre-trained conventional convolutional neural network ConvNeXt~\cite{liu2022convnet}, which is distinguished by its depth separable convolution architecture facilitating the disentanglement of pixel-level interactions and channel-wise information propagation, in this Section.

\subsection{Model Initialization}
\textbf{Initializing primary terms with existing weights:}
As depicted in Fig.\ref{fig:overview}(a), our methodology commences with the pre-trained convolutional neural network utilizing a large-scale dataset. Owing to the inherent characteristics of first-order computation, neurons within ConvNeXt execute the computation of the $W_CX+b$ segment as delineated in Eq.~\ref{equ:neu}. 

\textbf{Initializing quadratic terms to zeros:} At the beginning of the adaptation, we set the value of each element within $W_Q$ to 0 to ensure that it has the same output as the original network before adaptation. This step can be implemented more simply by initializing the $W_a$ matrix, described later in this Section, to 0 and $W_b$ with a Gaussian distribution.
\subsection{Tuning Conventional Neural Networks with Quadratic Adapter}
\textbf{Nonlinear Adaptation:}  During the adaptation stage, we maintain the immutability of these first-order parameters while redirecting our focus towards training a quadratic adaptation term $X^TW_QX$ (\textbf{Quadratic Adapter}) to modulate the model's output as follows:

\begin{equation}
y =  \underbrace{W_CX + b}_{\text{ConvNet}} + \underbrace{X^TW_QX}_{\text{Quadratic Adapter}}
\end{equation}

In adaptation stage, the introduction of such a Quadratic Adapter not only encapsulates linear distributional shifts in feature representation but also accommodates nonlinear quadratic shifts. Leveraging the low-rank decomposition technique for quadratic terms \cite{xu2023quadranet}, we effectively mitigate computational complexity from $O(n^2)$ to linear complexity $O(2kn)$. Therefore the quadratic neuron would be: 
\begin{equation}
y =  W_CX + b + X^TW_a^TW_bX
\label{equ:neurons}
\end{equation}

\textbf{Selective Adapter Placement:} The spatial high-order interactions between the pixels in the data point play a far more crucial role than the high-order interactions between the channels~\cite{xu2023quadranet}. However, inter-channel information interactions account for more than 90\% of computation in modern network design~\cite{liu2022convnet,xu2023quadranet,zhang2018shufflenet,howard2017mobilenets}. We therefore integrate the Quadratic Adapter term seamlessly with the less computationally burdensome depth-width convolution operator. Consequently, not only are the parameters and computational resources required for the quadratic adapter maintained at minimal levels but also the efficient modeling of high-order, nonlinear distributional shifts in pixel width is ensured.

\subsection{Inference with Library Optimized Quadratic Neural Networks}
\textbf{Library Acceleration:} As shown in Fig~\ref{fig:overview}, the integration of the quadratic adapter with the first-order terms of conventional neural networks yields a quadratic neuron within the convolutional layer. QuadraLib~\cite{xu2022quadralib} is instrumental in establishing a computational library, employing PyTorch-supported optimized conventional operators tailored for diverse types of Quadratic Neural Networks. By capitalizing on this acceleration capability, during the inference phase, we migrate the parameters of the quadratic adapter acquired during the tuning process to the quadratic terms of QuadraNet~\cite{xu2023quadranet}, like in Eq.\ref{equ:neurons}. This enables the quadratic model to fit the accelerated neuron format in QuadraLib~\cite{xu2022quadralib}.

%This enables the comprehensive model, fine-tuned to leverage quadratic neural networks, to achieve accelerated inference performance.
%quadratic adapter与常规的一次项结合，构成了一个在卷积层面的二次神经元。QuadraLib对各种类型的Quadratic Neural Networks使用了Pytorch支持的经过优化的常规算子建立计算库。我们利用这种加速特性，在inference阶段，我们的经过quadrtic kernel adapter的参数转移到QuadraNet的二次项中，使得tuning的模型整体使用二次神经网络进行加速的推理。

\textbf{Redundant Elimination:}
 Furthermore, in contrast to the vanilla QuadraNet~\cite{xu2023quadranet} block, we eliminate the residual connection following the Quadratic DW-Conv and omit layer normalization at the beginning of the block. This adjustment aligns the overall structure of the network with ConvNeXt, except for the depth-width convolution layers. This decision is rooted in the belief that the exchange of information between depth-width convolution and channel-width fully connected layers should be treated as a unified process of information interaction between two feature representation layers. However intermediate residual connections disrupt this process. Also, eliminating redundant residual connections allows intermediate states to be released more quickly during the inference stage, resulting in a halving of memory consumption during the inference phase. Additionally, it is believed that excessive normalization layers impair the model's generalization in modern learning theory~\cite{liu2022convnet}.

\section{Optimization}

\subsection{Efficient Atrous Quadratic Connection}

We identify a quadratic term whose computation can be further optimized. First, we reduce the number of connections in the quadratic adapter with an atrous connection. For a full quadratic connection, we have the output: 
\begin{equation}
    f_\text{full}(x) = \sum_{i,j}W_{a_i}W_{b_j}x_ix_j
\end{equation}
When generating feature $W_aX$ and $W_bX$, the atrous connection omits some of the input parameters to obtain a quadratic relation over a wider range with fewer interaction terms. 
This atrous connection is implemented by starting with one element in the weights marked as trainable, marking its neighboring elements as untrainable, and then selecting one of the remaining unassigned elements to be marked as trainable, and recursing in this manner until all the elements of the weights matrix are marked.
In an atrous connection, the output will be:
\begin{equation}
    f_\text{atrous}(x) = \sum_{i,j}\sum_{s \in \Omega_i}W_{a_s}x_i\cdot \sum_{t \in \Omega_j}W_{b_t}x_j,
\end{equation}
where $\Omega_i$ and $\Omega_j$ denote the neighborhood of i, j. 

\begin{figure}[t]
  \centering
  \includegraphics[width=\linewidth]{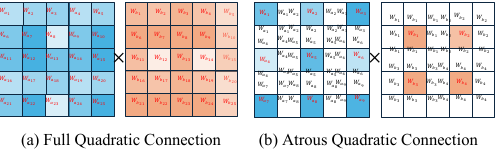}
  \caption{Atrous Quadratic Connections.}
  \label{fig:atrous}
  \vspace{-4mm}
\end{figure}

In atrous weight matrix, $W_a$ and $W_b$ are highly sparsified to reduce the number of total trainable parameters.
 As illustrated in Fig~\ref{fig:atrous}, a full quadratic connection term that incorporates 25 inputs has 2$\times$25 trainable parameters. While in the atrous quadratic connection, only those with red parameters are trainable, the rest weight is an arithmetic summation of its neighboring weights. Therefore, only 9 trainable parameters in $W_a$ and only 4 trainable parameters in $W_b$. Thus the number of parameters of a quadratic term is compressed from 50 to 13 in the example of Fig.~\ref{fig:atrous}. This design greatly reduces the number of parameters and computations.
%利用了类似空洞卷积减少connection数量，减少adapter的参数和计算量
It's worth noting that, when such operations are mapped to convolution, it's possible to leverage existing dilation convolution~\cite{yu2015multi} operators for high-speed performance optimization~\cite{paszke2019pytorch}.

%做一个进一步的分析

\subsection{Memory-Efficient Back-Propagation}

%反向传播过程的中间状态提前计算释放memory，并且高斯核中W_a和W_b的梯度中间状态相反，在这里可以只做一次计算，从而更新两组参数。
In neural network training, efficient memory management during the backpropagation process is crucial for handling large-scale networks. We propose an optimization strategy that significantly reduces the memory footprint by selectively retaining intermediate states essential for gradient computations in quadratic adapter's training. Specifically, we identify that the intermediate products $W_aX \odot W_bX$ and $W_cX$ can be released early in the computation, as these do not contribute to the dominant workload of subsequent operations. Furthermore, the gradients of weights $W_a$ and $W_b$ are dependent solely on the outputs of $W_bX$ and $W_aX$, respectively. This is due to the fact that:
\begin{equation}
	\begin{aligned}
	\frac{\partial \mathscr{L}}{\partial W_a^l} = &\frac{\partial \mathscr{L}}{\partial X^{l+1}}\cdot \overbrace{\frac{\partial X^{l+1}}{\partial (W_a^lX^l)(W_b^lX^l)}\cdot \frac{\partial (W_a^lX^l)(W_b^lX^l)}{\partial (W_a^lX^l)}}^{\text{Skip Intermediate Gradient}}\cdot \frac{\partial (W_a^lX^l)}{\partial W_a^l}\\
		= &\frac{\partial \mathscr{L}}{\partial X^{l+1}}\cdot 1 \cdot (W_b^lX^l)\cdot X^l,
	\end{aligned}
\label{equ:garident}
\end{equation}
where the second and third partial differentials equal to $1$ and $(W_b^lX^l)$, which means $W_aX \odot W_bX$ can be released immediately during front-propagation because it is not needed for back-propagation.
This allows us to optimize the memory usage further by retaining only the necessary intermediate outputs. This method ensures that only two intermediate states—$W_aX$ and $W_bX$—are retained, thus minimizing the memory requirements and computational costs. The implementation of this approach streamlines the computation and enhances the scalability of the training process in the computational resource extremely constrained environments.

%第10/11栏末尾

\section{Experiments}

\subsection{Experiment Settings}
\label{sec:train}
\textbf{Baseline:} We evaluate the performance and efficiency of QuadraNet V2 on ConvNeXT~\cite{liu2022convnet}. We create different variants ofQuadraNet V2, QuadraNet-T/S/B/L/XL based on the ConvNeXT-T/S/B/L/XL trained weights, to be of similar complexities. As with models such as ConvNeXT~\cite{he2016deep}, Swin~\cite{liu2022swin}, HorNet~\cite{rao2022hornet}, MogaNet~\cite{li2024moganet}, and QuadraNet~\cite{xu2023quadranet} we use ImageNet -1K~\cite{deng2009imagenet} as the experimental dataset, which
contains 1.28M training images and 50K validation images
from 1000 classes. Meanwhile, we list these models as a baseline.

\textbf{Training Details:} Since the vast majority of the linear term weight parameters in QuadraNet V2 have already been initialized. We only train a very small portion of the quadratic term (Quadratic Adapter) parameters distributed in the depth-width interaction, with 1/10 epochs as needed by the previous baseline models (300 epochs). We train QuadraNet V2 with the AdamW optimizer \cite{loshchilov2018decoupled} with an initial learning rate set to $4 \times 10^{-2}$. A weight decay parameter of 0.05 is applied.
The input images are processed at a resolution of $224 \times 224$ pixels for ImageNet-1K pre-trained model, and $224 \times 224$ or $384 \times 384$ for the adaptation of ImageNet-21K pre-trained models.
To enhance the stability of the training process, we implement gradient clipping, which constrains the maximum gradient value to 5. We use the ``Channel Last'' memory layout following~\cite{liu2022convnet} for better front-propagation efficiency. We use 4 NVIDIA A100s for model training.

\begin{table}
\small
  \caption{\textbf{ImageNet-1K Classification Results}}
  \label{tab:result}
  \centering
  \begin{tabular}{lllll}
    \toprule
%    \multicolumn{2}{c}{Part}                   \\
%    \cmidrule(r){1-2}
    Model     & Image     & Params& FLOPs& Top1 \\
         &Size     & (M)  & (G) & Acc.(\%) \\
    \midrule
        \multicolumn{4}{l}{\textit{ImageNet-1K Model Enhancing Adaptation}} \\
    \midrule
    ConvNeXt-T\cite{liu2022convnet} &224$^\text{2}$& 29 & 4.5 & 82.1\\
        HorNet-T\cite{rao2022hornet}&224$^\text{2}$ & 22 & 4  &82.8  \\
        MogaNet-T\cite{li2024moganet}&224$^\text{2}$ & 25 & 5.0  &83.5  \\
         QuadraNet-T\cite{xu2023quadranet} & 224$^\text{2}$& 23.6 & 4.1 &  82.2\\
    \rowcolor{mygray}\textbf{QuadraNet V2-T} & 224$^\text{2}$  & 29.1  & 4.5& 83.4  \\
\midrule
ConvNeXt-S\cite{liu2022convnet} &224$^\text{2}$& 50 & 8.7 & 83.1\\
        HorNet-S\cite{rao2022hornet}&224$^\text{2}$ & 50 & 8.8  &83.8  \\
        MogaNet-S\cite{li2024moganet}&224$^\text{2}$ & 44 & 9.9  &84.3  \\
         QuadraNet-S\cite{xu2023quadranet} & 224$^\text{2}$& 50.2 & 8.9 &  83.8\\
    \rowcolor{mygray}\textbf{QuadraNet V2-S}    & 224$^\text{2}$ & 51.3  & 8.7 & 84.3 \\
\midrule
    ConvNeXt-B\cite{liu2022convnet} &224$^\text{2}$& 89 & 15.4 & 83.8\\
        HorNet-B\cite{rao2022hornet}&224$^\text{2}$ & 87 & 15.6  &84.2  \\
        MogaNet-B\cite{li2024moganet}&224$^\text{2}$ & 83 & 15.9  &84.7  \\
         QuadraNet-B\cite{xu2023quadranet} & 224$^\text{2}$& 90.4 & 15.8 &  84.1\\
   \rowcolor{mygray}\textbf{QuadraNet V2-B}     & 224$^\text{2}$  & 90.1   & 15.4 & 84.8 \\
\midrule
   ConvNeXt-L\cite{liu2022convnet} &224$^\text{2}$& 198 & 34.4 & 84.3\\
   
    \rowcolor{mygray}\textbf{QuadraNet V2-L}  & 224$^\text{2}$  & 200   & 34.4 & 85.0 \\
     ConvNeXt-L\cite{liu2022convnet} &384$^\text{2}$& 198 & 101 & 85.5\\
    \rowcolor{mygray}\textbf{QuadraNet V2-L}  & 384$^\text{2}$  & 200   & 101.2 & 86.0 \\
    \midrule
    \multicolumn{5}{l}{\textit{ImageNet-21K Pretrained Model Adaptation or Fine-Tuning}} \\
    \midrule
    ConvNeXt-L*\cite{liu2022convnet} &224$^\text{2}$& 198 & 34.4 & 86.6\\
    MogaNet-L*\cite{li2024moganet} &224$^\text{2}$& 181 & 34.5 & 85.1\\
     HorNet-L*\cite{rao2022hornet}&224$^\text{2}$ & 195 & 34.8  &86.8  \\
    \rowcolor{mygray}\textbf{QuadraNet V2-L*}  & 224$^\text{2}$  & 200   & 34.4 & 87.4 \\
    \midrule
    ConvNeXt-XL*\cite{liu2022convnet} &224$^\text{2}$& 350 & 60.9 & 87.0\\
    \rowcolor{mygray}\textbf{QuadraNet V2-XL*}     & 224$^\text{2}$  & 350   & 61 & 88.3 \\
    ConvNeXt-XL*\cite{liu2022convnet} &384$^\text{2}$& 350 & 179 & 87.8\\
    \rowcolor{mygray}\textbf{QuadraNet V2-XL*}    & 384$^\text{2}$  & 350   & 180 & 89.3 \\
    \bottomrule
  \end{tabular}
\end{table}

\textbf{ImageNet Experiments:} On ImageNet, we design two sets of experiments, called Enhancing Adaptation and Pre-Training Transfer Adaptation. Enhancing Adaptation represents adapting a ConvNeXt model trained only by ImageNet-1K to QuadraNet V2 on the same ImageNet-1K dataset, which evaluates the Quadratic Adapter's inherent ability to model non-linearity in the original data for performance gain. Pre-Training Adaptation means initializing the   QuadraNet V2 with 14 million training sample's ImageNet-21K dataset pre-trained ConvNeXt, then adapting the model with Quadratic Adapter to distinct ImageNet-1K classification task. This evaluates the Quadratic Adapter's ability to learn the distribution shift between pre-training and downstream data.

\subsection{ImageNet-1K Enhancing Adaptation}
\textbf{State-of-the-art Performance:}
 In Table~\ref{tab:result}, the results of QuadraNet V2 models performance on ImageNet-1K are summarized. Compared to the original ConvNeXt~\cite{liu2022convnet}, which we used to initialize the primary term weights, the introduction of the secondary adapter resulted in performance gains on models of different sizes. The low-rank decomposition and Atrous Quadratic connection design allow further improvements in the modeling capacity of the quadratic terms compared to vanilla QuadraNet~\cite{xu2023quadranet}, leading to better classification performance of QuadraNet V2.

\textbf{Reduced GPU Time:} QuadraNet V2 maintains the state-of-the-art performance of the high-order approaches compared to other high-order models of the same size. But it is worth noting that the training time required for QuadraNet V2 is extremely compressed with the Quadratic Adapter technique. As shown in Table~\ref{tab:gpuhour}, using the training setting in Section~\ref{sec:train}, with Quadratic Adapter, the training GPU hours is only 7.7\% to reach the equal Top-1 accuracy of HorNet~\cite{rao2022hornet}.

\begin{table}
\small
  \caption{\textbf{Training Hours}}
  \label{tab:gpuhour}
  \centering
  \begin{tabular}{llll}
    \toprule
%    \multicolumn{2}{c}{Part}                   \\
%    \cmidrule(r){1-2}
    \multirow{2}{1em}{Model}     & Training  & Top1  & GPU \\
    &Strategy&Acc.(\%)&hours\\
    \midrule
        HorNet-T\cite{rao2022hornet}& from scratch &82.8 & 455 \\
         QuadraNet-T\cite{xu2023quadranet} & from scratch & 82.2 &377\\
             QuadraNet V2-T &  from scratch& 82.9  & 427\\
    \rowcolor{mygray}\textbf{QuadraNet V2-T} & Quad. Adapter& 82.9 & 35  \\

    \midrule
    \multicolumn{2}{l}{
    \textit{ImageNet-21K Pre-trained Model}}\\
    \midrule
     HorNet-L*\cite{rao2022hornet}& from scratch  & 86.8 & 19270 \\
     QuadraNet V2-L*  & from scratch  &  87.3 & 16203\\
   \rowcolor{mygray} \textbf{QuadraNet V2-L*}  & Quad. Adapter  &  87.4 & 260\\

    \bottomrule
  \end{tabular}
\end{table}

\subsection{ImageNet-21K Pre-Training Adaptation} 

\textbf{Performance Gains with Quadratic Adapter:} In Table~\ref{tab:result}, we evaluate the effectiveness of QuadraNet V2 on larger scale models under ImageNet-21K pre-training. With a -L scale model of about 200M parameter size, we find that QuadraNet V2 achieves performance that comprehensively outperforms other high-order models. This is due to the fact that the non-linearity of the data distributional shifts is all modeled into the nonlinear pattern of the quadratic terms of the quadratic neural network, instead of the linear pattern of the traditional model or the hybrid pattern of other high-order networks. This performance gain is enhanced even more in the larger scale -XL models. The Quadratic Adapter in QuadraNet V2-XL brings a remarkable 1.3\% Top1 Accuracy to the model, compared to the original full fine-tuning.

\textbf{Extremely Reduced GPU Time:} In terms of total GPU hours required for training a large-scale model, the advantage of QuadraNet V2 is even more pronounced in this "Pre-Training + Adaptation" paradigm, because our model omits the large amount of GPU time required for pre-training due to the direct use of existing pre-trained weights to initialize the model. As shown in Table~\ref{tab:gpuhour}, this results in saving \textbf{98.6\%} of training GPU hours while achieving stronger performance for our model than building a HorNet-XL from scratch.

\subsection{Comparison with Tuning Techniques}
\begin{table}
\small
  \caption{\textbf{Comparison of Adaptation Methods}}
  \label{tab:adapt}
  \centering
  \begin{tabular}{llll}
    \toprule
%    \multicolumn{2}{c}{Part}                   \\
%    \cmidrule(r){1-2}
    \multirow{2}{1em}{Model}     & Adaptation  & Top1  & GPU \\
    &Strategy&Acc.(\%)&hours\\
    \midrule
     ConNeXt-L*\cite{liu2022convnet}& full fine-tuning  & 86.6 &  447\\
     ConNeXt-L*\cite{liu2022convnet}& LoRA~\cite{hu2021lora}  & 84.7 & 257 \\
   \rowcolor{mygray} \textbf{QuadraNet V2-L*}  & Quad. Adapter  &  87.4 & 260\\

    \bottomrule
  \end{tabular}
\end{table}
In Table~\ref{tab:adapt}, we show the comparison with full fine-tuning and LoRA~\cite{hu2021lora}. Compared to LoRA, QuadraNet V2 achieves stronger performance with almost the same adaptation overhead due to its stronger modeling ability for nonlinear shifts between pre-training data and task data. Compared to full fine-tuning, the QuadraNet V2 saves about 42\% GPU times while achieving 2.7\% higher accuracy. This reveals that the high-order neural network constructed by Quadratic Adapter has a stronger model capacity than the traditional network.

\subsection{Limited Budget Training}

We uniformly set an upper limit of 275 GPU hours of training to evaluate the training accuracy that can be achieved with the same budget by using the QuadraNet V2 framework and the traditional "Pre-Training + Fine-Tuning" paradigm to train a large size high-order model. As shown in Table~\ref{tab:budget}, only QuadraNet V2 can make a large high-order model reach its optimality with extremely limited training. In this case of extremely limited computational budgets, a large, high-order network constructed with QuadraNet V2 can achieve an accuracy improvement of about 23.8\% to 34.2\%. To the best of our knowledge, our framework is currently the only training method that can make a performant large high-order model converge to optimal at this training budget.

Meanwhile, we observe from the baseline experiments of HorNet and MogaNet that, with a small number of training epochs, 1 epoch of pre-training on large-scale data is inferior to directly using the corresponding resources on the target.
\begin{table}
\small
  \caption{\textbf{Comparsion of High-Order Models' Performance with Limited Training Budget}}
  \label{tab:budget}
  \centering
  \begin{tabular}{lccc}
    \toprule
%    \multicolumn{2}{c}{Part}                   \\
%    \cmidrule(r){1-2}
Model& GPU hours & Epochs(Pre.+Adap.) & Acc.(\%)\\
\midrule
HorNet-L~\cite{rao2022hornet} & 289 & 1+20 & 57.7\\
HorNet-L~\cite{rao2022hornet} & 274 & 0+30 & 63.6\\
MogaNet-L~\cite{li2024moganet} & 272 & 1+15 & 53.2\\
MogaNet-L~\cite{li2024moganet} & 263 & 0+25 & 59.8\\
 \rowcolor{mygray}\textbf{QuadraNet V2-L*} & 260 & 0+30 & 87.4\\
    \bottomrule
  \end{tabular}
\end{table}
\subsection{Downstream Tasks Transfer Capacity}

\begin{table}
  \caption{\textbf{MS COCO Detection Results}}
  \label{tab:detection}
  \centering
  \begin{tabular}{lllcc}
    \toprule
%    \multicolumn{2}{c}{Part}                   \\
%    \cmidrule(r){1-2}
    Model     & AP$^\text{box}$  & AP$^\text{mask}$    & Params  & FLOPs)  \\
    \midrule
    ConvNeXt-T~\cite{liu2022convnet} & 50.4  & 43.7& 86M  & 741G \\
    HorNet-T~\cite{rao2022hornet} & 51.7  & 44.8& 80M  & 730G\\
   \rowcolor{mygray} \textbf{QuadraNet V2-T} & 52.0  & 45.6& 86M  & 742G \\
    ConvNeXt-S~\cite{liu2022convnet} & 51.9 & 45.0  & 108M&  827G \\
    HorNet-S~\cite{rao2022hornet} & 52.7  & 45.2& 107M & 830G \\
   \rowcolor{mygray} \textbf{ QuadraNet V2-S }   & 53.0 & 45.3  & 109M&  828G \\
    ConvNeXt-B~\cite{liu2022convnet} & 52.7  & 45.6& 146M & 964G \\
    HorNet-B~\cite{rao2022hornet} & 53.3  & 46.1& 144M  & 969G \\
   \rowcolor{mygray} \textbf{ QuadraNet V2-B }    & 53.8 &  46.7& 147M& 966G \\
       \midrule
    \multicolumn{5}{l}{\textit{ImageNet-21K Pretrained Backbone}} \\
    \midrule
        ConvNeXt-L*~\cite{liu2022convnet} & 54.8  & 47.6& 255M & 1354G \\
    HorNet-L*~\cite{rao2022hornet} & 55.4  & 48.0& 251M  & 1363G \\
   \rowcolor{mygray} \textbf{ QuadraNet V2-L* }    & 55.8 &  48.2& 258M& 1355G \\
    \bottomrule
  \end{tabular}
  \vspace{-6mm}
\end{table}

Further, we verified the ability of object detection~\cite{zou2023object}, a downstream task, of QuadraNet V2 on the MS COCO~\cite{lin2014microsoft} dataset. 
We adopt the Cascade Mask R-CNN~\cite{cai2018cascade,cai2019cascade} framework for object detection with backbone network substituted as QuadraNet V2-T/S/B. In Cascade Mask R-CNN, following baseline ConvNeXt~\cite{liu2022convnet} and HorNet~\cite{rao2022hornet}, we use a 3$\times$ schedule. 
Unlike other baseline imported ImageNet-1K trained backbone models that are fully fine-tuned during the object detection training process, we still use the Quadratic Adapter for backbone network Adaptation.
That means QuadraNet V2's primary term weights remain frozen, and only the parameters in the Quadratic Adapter can be updated. In Table~\ref{tab:detection}, we report the evaluation of box AP and mask AP~\cite{lin2014microsoft} of our models and ConvNeXts and HorNets. Our models significantly outperform ConvNeXt family models in the box and mask AP at the same scale, suggesting the advantage of the Quadratic Adapter in modeling the nonlinearity of the data shift.
Our model achieved a slightly higher AP than HorNet. However, it is worth noting that our backbone models still omit the pre-training process, which greatly saves the total computational overhead of training the detection models.

\section{Conclusion}

In this work, we present QuadraNet V2 --- a revolutionary model training framework for the efficient and sustainable development of high-performance models. By using quadratic neuron, we reduce the granularity of the high-order model from the model architecture level to the neuron level. Based on this, QuadraNet V2 avoids the tremendous computational overhead of pre-training by leveraging existing legacy pre-trained weights to build high-order neural with a Quadratic Adapter. Further, we perform low-rank and atrous optimizations on Quadratic Adapters to gain performance and computational advantages. Most importantly, these aforementioned properties of QuadraNet V2 point the way to building new high-order models more efficiently and sustainably in the trend toward large-scale, high-performance models.

\balance
\bibliographystyle{_sty/ACM-Reference-Format}
\bibliography{_bib/sample-sigconf}

\begin{comment}
\appendix

\section{Research Methods}

\subsection{Part One}

Lorem ipsum dolor sit amet, consectetur adipiscing elit. Morbi
malesuada, quam in pulvinar varius, metus nunc fermentum urna, id
sollicitudin purus odio sit amet enim. Aliquam ullamcorper eu ipsum
vel mollis. Curabitur quis dictum nisl. Phasellus vel semper risus, et
lacinia dolor. Integer ultricies commodo sem nec semper.

\subsection{Part Two}

Etiam commodo feugiat nisl pulvinar pellentesque. Etiam auctor sodales
ligula, non varius nibh pulvinar semper. Suspendisse nec lectus non
ipsum convallis congue hendrerit vitae sapien. Donec at laoreet
eros. Vivamus non purus placerat, scelerisque diam eu, cursus
ante. Etiam aliquam tortor auctor efficitur mattis.

\section{Online Resources}

Nam id fermentum dui. Suspendisse sagittis tortor a nulla mollis, in
pulvinar ex pretium. Sed interdum orci quis metus euismod, et sagittis
enim maximus. Vestibulum gravida massa ut felis suscipit
congue. Quisque mattis elit a risus ultrices commodo venenatis eget
dui. Etiam sagittis eleifend elementum.

Nam interdum magna at lectus dignissim, ac dignissim lorem
rhoncus. Maecenas eu arcu ac neque placerat aliquam. Nunc pulvinar
massa et mattis lacinia.
\end{comment}
\end{document}